# Semantic Description of Parameters in Web Service Annotations

Jochen Gruber[1]

**Abstract.** A modification of OWL-S regarding parameter description is proposed. It is strictly based on Description Logic. In addition to class description of parameters it also allows the modelling of relations between parameters and the precise description of the size of data to be supplied to a service. In particular, it solves two major issues identified within current proposals for a Semantic Web Service annotation standard.

## 1 INTRODUCTION

The vision of the Semantic Web, [1], is to make the large amount of data on the Web machine-processable. This shall be achieved by semantically annotating web resources, i.e. elements of the resources are associated with elements of ontologies. Under the auspices of the W3C, a couple of language standards have been developed and widely accepted: building on the Resource Description Framework, RDF, [11], and the RDF Schema language, [12], concepts and concept relations can be defined in the Web Ontology Language OWL, [7].
However, the web also offers access to services, i.e., Web Services. For Web Services, there currently exist standards as SOAP, [14], WSDL, [19], and UDDI, [16]. These standards operate at a purely syntactic level as they are intended to free programmers from technical details of connecting with services. The goal of Semantic Web Service technology is the combination of Semantic Web technology and Web Services. By using ontologies as a semantic data model for Web Service technologies, Web Services have machine-processable annotations just as static data on the Web.
The above mentioned Semantic Web language standards have been developed for describing static web resources and are based on Description Logic (DL), [2]. While Description Logic is far less expressive than, say Horn Logic, there exist very efficient implementations for processing corresponding data bases. However, it is clear that many descriptions of web resources require more expressive languages and consequently, there are intensive ongoing discussions on extensions of OWL by rules as suggested in the Semantic Web Rule Language proposal, SWRL, [15], the Rule Markup Language RuleML, [13] and the Web Service Modeling Language WSML, [22]. Indeed, much of the current work on Web Service annotation relies on rules: The Web Service Modeling Ontology, WSMO, is based on WSML and version 1.1 of OWL for services, OWL-S, on SWRL.

The following discussion shall be illustrated by a typical use case: The *StockQuote* service is publicly available from the WebserviceX web site, [18]. There, one can find a detailed description of the service including its WSDL file and a test page. The service expects as input a string containing a stock ticker symbol and returns the current trading price of the corresponding stock at the New York Stock exchange, for more details see [18].
Semantically, the input string can be modelled by some instance of a class `TickerSymbol`, while the output is the current stock price of type `xsd:double`. It could be related to the input parameter via some relation `hasPrice`. However, it is clear that this relation can not simply be a data type relation with range of type `xsd:double`: every price consists of the currency it is related to and its numeric value. Therefore, as a semantic service it should return an instance of some class, say `MonetaryValue`, with some object property `hasCurrency` which has as range instances of a class `Currency` and some data type property `numericalValue` of type `xsd:double`. In particular, now the service annotation describes that the return value is meant in US Dollar, a fact which is not even mentioned in the WSDL file and thus can be derived only by a human programmer from the fact that the stock quotes are from the New York Stock exchange, again not mentioned in the WSDL file. Even if these facts would be mentioned in some documen-tation part of the WSDL file, it clearly could not be processed perfectly. Thus, we see that the semantic description generally contains elements not contained in the syntactic description of the service.
In graph notation we get the following simplified ontology in figure 1:

---

[1] SAP AG, Germany, email: jochen.gruber@sap.com

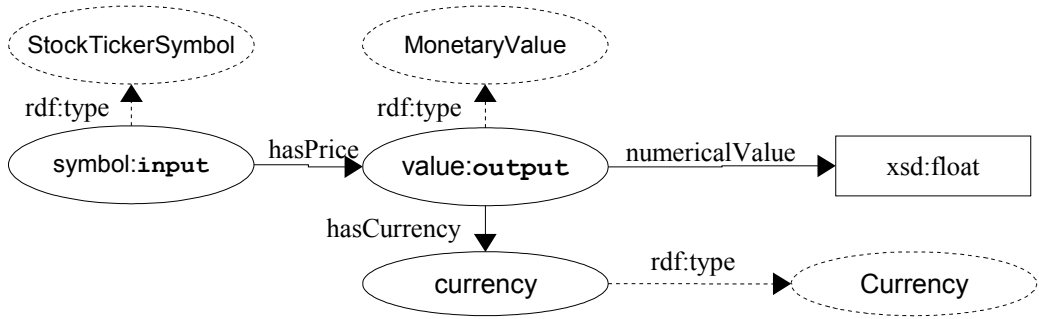

**Figure 1.** Ontology for StockQuote service

In section two we have a quick overview on current proposals for Web Service annotations. Focusing on parameter descriptions, we discuss the pro and cons of current proposals and identify potential problems in practical use cases. In section three an alternative description is proposed and compared to the approaches presented in section two. It is also shown how this approach solves the identified potential problems. Section four concludes this work. In section four a programming framework four automated selection, composition and invocation based on the introduced annotation is presented.

## 2 ANNOTATIONS

Currently, no annotation standard for Web Services exists. However, there exist a range of proposals for such a standard, the most prominent ones are:
- OWL for services, OWL-S, in versions 1.0, [9], and 1.1, [10], a collaborative effort of academic and industrial partners within the DARPA Agent Markup Language pro-gram.
- WSDL for services, WSDL-S, [20], a specification by IBM and the METEOR group, [6], at the University of Georgia.
- Web Service Modeling Ontology, WSMO, [23], an extension and refinement of the Web Service Modeling Framework, WSMF, [21].

It is not our goal to develop a new framework for service annotation, rather, OWL-S is modified for solving two critical issues identified below. However, by considering the other two proposals, one can see easily that corresponding modifications could be applied there as well. This is a strong indication reason for the annotation used here to be accommodative to some future W3C annotation standard for Semantic Web Services.

In all three annotation frameworks one can identify a common set of main annotation tasks:
- Non-functional properties like author and contact information.
- Service API, describing the service parameters and the service's effect on them. This is done by association of semantic parameters with elements in some ontology, i.e. with classes as in OWL-S version 1.0 and WSDL-S or with variables as in OWL-S version 1.1 and WSMO.
- Service orchestration, describing how the calls to individual operations of a service have to be related for yielding a successful service invocation.
- Real world conditions and effects of service invocation.
- Mediation between semantic and syntactic description of a service, the latter being represented by the service's WSDL file.

The non-functional properties are not critical. The mediation between semantic and syntactic description is specified largely by the decision on the semantic description. The orchestration of services and their real world conditions and effects are not considered here as these are currently the most intensively researched topics of Web Service annotation and beyond the scope of this work. Moreover, one can identify a large segment of services where these aspects are irrelevant as in [4].

We are thus left to consider the semantic annotation of parameters: It not only has to describe the semantics of the individual parameters, but also their relation, particularly relations between parameters of different messages. For sake of simplicity we focus on the input and output of a single operation, however where necessary it will be emphasized that the considerations also apply to more complex scenarios.

In all three frameworks listed above, ontologies are used to semantically describe service para-meters, either by directly associating them with classes or by associating them with variables which in turn can be restricted to certain classes. This can be done by referring to some ontology defined in a DL language as in the case of OWL-S version 1.0 which is restricted to ontologies described in OWL. In case variables are used, obviously an ontology language is required which includes the definition of rules as in OWL-S version 1.1 which is restricted to ontologies described in SWRL. In the following, we call these ontologies the semantic models of the services.

## 2.1 Problems

In [9], section five, there is noted a problem regarding identification of parameters of different messages. This problem is clearly solved by association of parameters with variables a done in [10] and [22]. Variables can be referred to by several parameters, implicitly identifying those which re-fer to the same one. However, the problem of identification of parameters is only a facet of a more general problem with the approach of directly associating parameters with classes. This problem can be formulated as follows: For describing how an operation processes its input parameters to yield its output parameters, one has to describe how these parameters are related with each other. While one has generally in an ontology many relations with domains and ranges including the service parameter types, not all these relations will hold for the class instances with which the parameter is called and which are returned by it.

Clearly, employing rules and variables, one can describe these relations. However, the use of rules has its own drawbacks:

- First, it requires a language beyond DL, implying worse performance and higher complexity.
- Second, either potential ontology authors are required to have knowledge of the intricacies of logic programming, or they have to rely on authoring tools hiding the actual rules by generating them e.g. from descriptive models, necessarily limiting the range of rules to be used.

Take as example the StockQuote service: The returned `MonetaryValue` instance is not an arbitrary one, rather it is the one corresponding to the supplied `TickerSymbol` instance, the two being related by property `hasPrice`. While for a human reader this may be clear, this information is not contained explicitly in an annotation associating parameters with ontology classes. The only way the framework could proceed is by inferring that for all input and output parameters whose associated semantic classes have relations, these relations also hold for the instances. While this may be a sensible heuristic easily implemented by such a framework, it can not be the final solution: One can easily find examples, where there are several relations between concepts associated with parameters, even pairwise exclusive ones: take a variant of the StockQuote service which returns three instances instead of one `MonetaryValue`, namely one for the currently highest bit, one for the currently lowest offer and one for the last trade price of the stock (indeed, these three prices are usually returned by most stock information services). Correspondingly, there are three relations between `TickerSymbol` and `MonetaryValue`, e.g. `lastTrade`, `bit` and `ask`.

Obviously, if we associate parameters with variables, one can add rules describing the relations. However, such rules would be just facts about the validity of relation. The modeling of facts however is precisely the domain of DL. Generally, modeling within DL is more intuitive than within Functional Logics, thus a solution for this problem within the bounds of DL would be desirable.

The semantic model is also used to describe which information from the knowledge base is used by the actual service: in general, semantic and syntactic parameters are not one to one, as seen in the StockQuote service. In this case, some mediator hast to be supplied with a certain fragment of the data base to be transformed into the syntactic parameter. In other words: an instance of the class, a parameter is associated with, has to be serialized and drawn from the data base. From object oriented programming it is well known that serialization is a non-trivial issue and we will show in the following how this issue materializes in Semantic Web Services processing.

There are two alternatives strategies for generic serialization:

- Supply only the input parameter instance, which in general is only a URI, and its associated datatype relations, which may miss large parts of the required data.
- Supply the complete subgraph reachable from the input parameter, which in case of the existence of cycles within the graph can become considerably more data then needed.

For illustration, take the following simple use case: a database on addresses, consisting of four classes `Address`, `Person`, `SurName` and `FamilyName`, four object relations `hasPerson` relating an `Address` to `Persons`, `hasAddress` relating a `Person` to its `Address`, `hasSurName` and `hasFamilyName` relating a person to its respective names and two datatype relations relating the `SurName` and `FamilyName` to their respective string presentations.
A fragment of the database including two persons and their address is shown in figure 2 (instances are enumerated by their class names in lower type letters, not all relations are named):

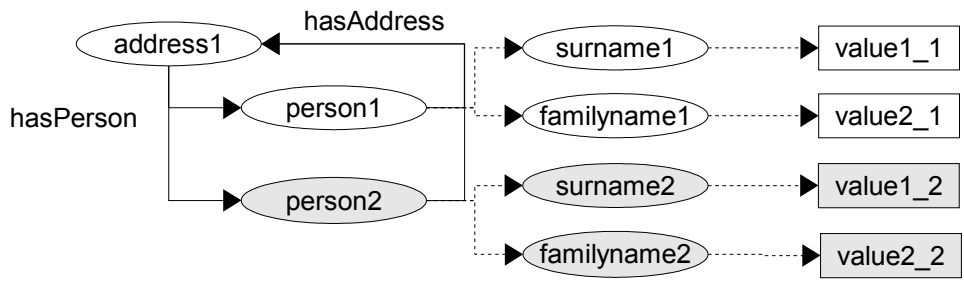

**Figure 2.** Database fragment for address database

Consider now a service asking for the name of a person: Its semantic input parameter is of type `Person`, while the syntactic parameter consists of the concatenated strings of both surname and family name. In particular, a mediator between semantic and syntactic parameters takes the whole grayed subgraph. Recalling the two generic strategies for data extraction, the first one supplies not enough information (the `Person` class does not have any datatype relations), while the second one supplies the whole class, including two instances of `Person`, where only one is expected.

To avoid this problem, the service consumer has to know precisely which information to supply to the service. However, contrary to object oriented programming, it is not a task of the ontology engineer to define a serialization of his classes: As reuse of ontologies is essential to the success of the Semantic Web, general ontologies won't be designed for specific applications, it depends on the mediator which fragment of the data base it precisely needs and therefore this information must be contained in the service annotation.
None of the above listed annotation frameworks do say anything about this problem. Clearly, with the help of logic expressions one can describe which information has to be supplied to the transformation. However, such a description would do nothing else than extract triples from the semantic database. Thus, it would simply be a recipe for extracting the precise fragment of the data base to be supplied for transformation. However, such a fragment is nothing else than an A-box.
Therefore, a solution using an A-box consisting of instances of the semantic model is presented in the next section. Such a solution is preferable not only because of theoretical reasons but also because of practical reasons: For a service author, particularly once semantic modelling has become part of main stream programming, defining an instance model with the help of some graphical development environment is much more intuitive than defining logical expressions.

## 3 INSTANCE MODELS

To solve the issues discussed in section 2.1, so called instance models consisting of A-boxes referring in turn to the semantic model are suggested. The idea is to model explicitly the situations where a user invokes a service: What does it mean to invocate a service with data taken from a semantic database and to insert the result back into the database?
a) We have to take certain triples from the database, supply them to the service and in turn insert new triples generated by the service back into the database. In particular, the condition on the database to invocate the service is that a certain fragment of defined triples has to exist in the database. This fragment itself is an A-box which we model by providing a template of the fragment expected by the service, we call it the input model.
b) The triples returned by the service again form an A-box, the output model. Of particular importance is the fact that some of the instances of the input model als occur in the output model, thus defining the precise relation between these parameters.
Three two instance models contain an instance for every semantic service parameter, in addition, the instance model contain all elements which have to be supplied to the service. E.g., in OWL-S the process model would point to these instances instead of the type definitions or the variables, respectively, in the semantic model. The type in turn can be derived from the instance model, so we do not need pointers to the types anymore.
The use of an instance model thus affects only the `ServiceModel` class in OWL-S, more pre-cisely the parameter definition. Instead of pointing to classes defined in some ontology as done in OWL-S 1.0, or to variables in some ontologies as done in OWL-S 1.1, semantic parameters are defined by pointing to some instance of an A-box, which in turn builds upon the semantic model.
For illustration, the StockQuote service has the output model shown in figure 3 (the input model is rather trivial):

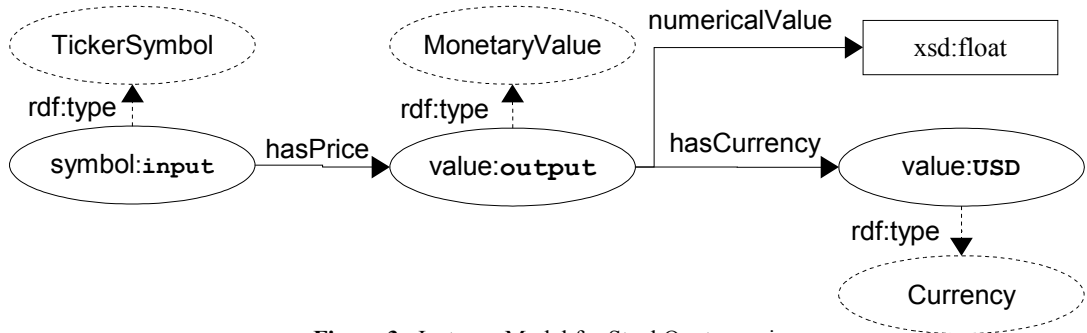

**Figure 3.** Instance Model for StockQuote service

The modelled relations (that is the relations which are explicitly included in the model as triples) are drawn with continuous lines, implicit relations (that is relations which can be inferred from the semantic model) by pointed lines.

This model in fact describes precisely the relations which hold between input and output parameters. Regarding data supply, this is no issue in the StockQuote case. However, it is clear that relations pointing to elements of the database not to be supplied to the services simply would be left out of the instance model. Thus, in our address database example, the input model would be an A-box corresponding to the grayed subgraph, leaving out the `hasAddress` relation, thus avoiding the supply of too much data.

Obviously, the instance models can also be employed in more complex scenarios, that is in scenarios where more than one process interact. The semantic descriptions of parameters of several processes can point to the same instance, other parameters of the same semantic type can point to different instances, thus describing which parameters of the same type refer to the same instance in the semantic data base and which to different ones.

## 4 A PROCESSING FRAMEWORK

The following approach for processing the presented annotation aims at tightly coupling an OWL data base and Information Providing Services. To do so the service is represented by a set of horn rules, describing the required input parameters as conditions for rule execution and the output parameters relations with the input parameters as the consequences of the rules. So if some rule engines queries on the data base, service invocation is triggered by resolution. In particular, the rule engine automatically selects the rules whose consequences contain the information needed by the query. Moreover, resolution also provides the means to compose services (in the required order) in case that no single service can provide the needed information.

Now OWL itself does not allow definition of additional rules not contained in the language standard. However, rules are already used in actual processing frameworks. For the realization we employ one for the Java language, the Jena framework by Hewlett Packard, [5], an open source software package under HP copyright but freely distributable and widely used within the Semantic Web community.

The implementation is designed as an add-on, building on Jena without modifying the Jena code itself. Clearly such an approach has the drawback of potential overhead which could be avoided by a deeper integration. On the other hand it has the advantage of far easier portability in case someone wants to use another underlying framework with similar capabilities – in case one wants to use a proprietary software package, deep integration may not be possible at all. Moreover, Jena allows the addition of other inference engines, while a deep integration would affect the engine.

Nevertheless, a main goal of this implementation is to tightly couple services and the knowledge base in such a way that some application built on Jena can add it without affecting Jena's runtime behavior, thus, the service calls become transparent to the using application (transparency only with regard to programming *logics*: obviously, processing time is drastically affected by service calls as the latter take much more time than simple rule invocation on orders of magnitude).

The reasoner included in Jena uses two rule engines, one forward and one backward chaining rule engine, the latter one based on the RETE algorithm, [3]. We do not want to go into the details of the rule syntax used, the interested reader can get this information from the documentation at the Jena homepage, [5], where the reasoner is described under 'Jena inference support'.

In the following we will use a notation for rules in the form $r: b_1, \ldots, b_n \rightarrow h_1, \ldots, h_m$ meaning that the body of the rule $r$ consists of a conjunction of terms $b_i$ which are the condi-tions under which the rule can be fired, and the head of $r$ consists of a conjunction of terms $h_i$ which are the consequences of $r$. This fits nicely with the instance models: the conjunction of triples in the input model corresponds to the rule body, while the head corresponds to the conjunction of triples in the output model. Further considerations lead us to remove those triples not containing any instances from the instance model as these triples would cause invocation of the rule with unrelated input parameters. In particular, no type definitions are contained in the head (contrary to the body, where they play an essential role).

Suppose a query is fired which asks for a triple as described in the head of a rule, resolution will instantiate the variables in the head. If and only if the variables corresponding to input parameters are instantiated such that they satisfy the rule's body term up to the service build-in, the latter will also be evaluated. In particular, all input variables are be bound to instances of the types of their corresponding service input parameters when the build-in is called. This means, the service is automatically selected and invoked when the information it provides is needed by a data base query.

Now suppose several services' rules are added to the rule engine. Then one might need information provided by service $A$ to call service $B$. However, what does it mean that invocation of $A$ requires data returned by $B$? In this case, there is some body term in $B$'s rule which describes some relation between instances (bound input variables) to hold if service $B$ is to be called and which is a (sub-) relation of elements returned by $A$. During resolution of $B$'s body thus $A$'s rule is fired, corresponding to invocation of service $A$.

We see therefore that not only automated selection and invocation, but also automated service composition is achieved by the mechanism of resolution.

## 5 RELATED WORK

Three of the most important suggestions for Semantic Web Service annotations have been presented ins section 2, the Web Service Modeling Framework, WSMF, [21] also includes a complete processing framework. There are also a range of other projects with a more limited scope, similar to the one presented in section 4: SWORD, [8] is a planning tool for generating a composite service out of a given set of basic services, given a description of the resulting service's input and output. It does not have any integration with Semantic Web data bases. The Web Service Description Framework, [17], is an approach analogous to the one presented here, however aiming at integrating Web Services with relational data bases. In particular, inferencing is limited to the capabilities provided by SQL.

## 6 CONCLUSION AND FUTURE WORK

A modification of the OWL-S Web Service annotation framework was presented, which allows the description of parameter relations and the precise determination of data to be extracted from the semantic database for service invocation. It allows to keep OWL-S description of service para-meters to be strictly kept within the domain of Description Logic and thus to maintain the good computability behavior of DL. Based on this description, a processing framework for tightly coupling information providing services and semantic databases. It allows to integrate Semantic Web Services transparently to an application building on the database and automates service selection, composition and invocation.

For future work the most critical issue is rule interaction: the more services are integrated, the more rules are added to the rule engine. With extension of ontology languages by rules, this number still increases. As it is well known from logic programming, e.g. as experienced in PROLOG, resolution results heavily depend on the order the rules are processed, easily leading to infinite resolution paths without careful rule base design. In scenarios with rules being defined in external ontologies and service processing frameworks, new ideas on modularizing the rule base for avoiding these problems are of highest importance. However, it is clear that this is a general problem, arising as soon as the rule base can be extended by the ontology engineer.